\documentclass[11pt]{article}

\usepackage[preprint]{acl}

\usepackage{times}
\usepackage{latexsym}
\usepackage{multirow}
\usepackage[table]{xcolor}
\usepackage{booktabs}
\usepackage{array}
\usepackage{subcaption}
\usepackage{tcolorbox}   
\usepackage{xcolor}    
\usepackage{fontawesome5}
\usepackage{hyperref}
\usepackage[T1]{fontenc}
\usepackage[utf8]{inputenc}
\usepackage{microtype}
\usepackage{inconsolata}
\usepackage{graphicx}
\usepackage{amsmath}
\usepackage{amssymb}
\usepackage{mathtools}
\usepackage{amsthm}

\usepackage[capitalize,noabbrev]{cleveref}
\usepackage{lipsum}
\theoremstyle{plain}
\newtheorem{theorem}{Theorem}[section]

\theoremstyle{definition}
\newtheorem{definition}[theorem]{Definition}

\theoremstyle{remark}

\newcommand{\ie}{\textit{i}.\textit{e}.}
\newcommand{\eg}{\textit{e}.\textit{g}.}
\title{VAUQ: Vision-Aware Uncertainty Quantification for LVLM Self-Evaluation}

\author{
 \textbf{Seongheon Park\textsuperscript{1}}\quad
 \textbf{Changdae Oh\textsuperscript{1}}\quad
 \textbf{Hyeong Kyu Choi\textsuperscript{1}} \\
 \textbf{Sean Du\textsuperscript{2}}\quad
 \textbf{Sharon Li\textsuperscript{1}}
\\
 \textsuperscript{1}University of Wisconsin-Madison \quad
 \textsuperscript{2}Nanyang Technological University \\
  \texttt{\{seongheon\_park, sharonli\}@cs.wisc.edu} }

\begin{document}
\maketitle
\begin{abstract}
Large Vision-Language Models (LVLMs) frequently hallucinate, limiting their safe deployment in real-world applications.
Existing LLM self-evaluation methods rely on a model’s ability to estimate the correctness of its own outputs, which can improve deployment reliability; however, they depend heavily on language priors and are therefore ill-suited for evaluating vision-conditioned predictions.
We propose \textbf{VAUQ}, a vision-aware uncertainty quantification framework for LVLM self-evaluation that explicitly measures how strongly a model’s output depends on visual evidence.
VAUQ introduces the Image-Information Score (IS), which captures the reduction in predictive uncertainty attributable to visual input, and an unsupervised core-region masking strategy that amplifies the influence of salient regions.
Combining predictive entropy with this core-masked IS yields a training-free scoring function that reliably reflects answer correctness.
Comprehensive experiments show that VAUQ consistently outperforms existing self-evaluation methods across multiple datasets\footnote{\url{https://github.com/deeplearning-wisc/vauq}}.

\end{abstract}

\section{Introduction}

LVLMs have demonstrated remarkable progress across a wide range of multimodal tasks, exhibiting strong visual understanding capabilities.
However, LVLMs remain prone to hallucinations, posing significant risks in high-stakes domains~\cite{liu2024survey}.
To assess model outputs, many existing works rely on external evaluators and  judges~\cite{liu2024mitigating,lee2024prometheus}.
Nevertheless, this approach is both costly and susceptible to hallucinations from the evaluator itself~\cite{xu2024hallucination}.

A promising direction for improving deployment reliability is \emph{self-evaluation}, in which a model estimates the correctness of its own outputs using internal signals, without relying on external supervision or auxiliary models. In language-only settings, prior work has shown that uncertainty quantification, consistency checks, and latent-state analysis can enable large language models to identify unreliable generations of themselves~\cite{malinin2020uncertainty, manakul2023selfcheckgpt, kuhnsemantic, du2024haloscope, orgad2024llms, wang2024latent}.
However, extending these self-evaluation techniques to LVLMs is non-trivial. Unlike text-only models, LVLMs operate over heterogeneous modalities, and their uncertainty arises from both linguistic and visual sources.

\begin{figure}[t]
    \centering
    \includegraphics[width=0.9\linewidth]{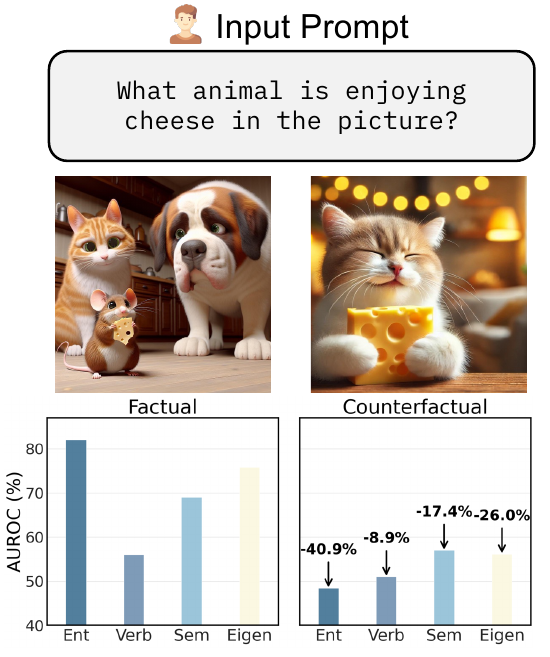}
    \caption{
        \textbf{Failure of LLM-based self-evaluation under language prior dominance.} Methods include: Entropy (Ent), Verbalized Confidence (Verb), Semantic Entropy (Sem), and EigenScore (Eigen). Performance comparison on the ViLP dataset using LLaVA-1.5-7B, which contains paired factual and counterfactual images associated with the same prompt. 
        Common self-evaluation methods often fail in counterfactual samples.
    }
    \vspace{-5mm}
    \label{fig:2}
\end{figure}

In practice, LVLMs can exhibit a strong language-prior dominance, relying heavily on statistical regularities learned during large-scale language pretraining while under-utilizing visual evidence~\cite{liu2024paying, leng2024mitigating, lee2025vlind, long2025understanding}. As a result, existing LLM-based self-evaluation methods can assign low uncertainty to hallucinated responses when the image contradicts common linguistic expectations (see~\cref{fig:2}). 
In such cases, confidence reflects fluency rather than grounding. 
This fundamental mismatch highlights the need for an approach that explicitly accounts for how much visual information contributes to a model’s prediction.

To this end, we introduce \textbf{V}ision-\textbf{A}ware \textbf{U}ncertainty \textbf{Q}uantification (\textbf{VAUQ}), a training-free framework for LVLM self-evaluation that measures whether a model’s confidence is justified by visual grounding. Our central insight is that informative and correctly utilized visual evidence should reduce predictive uncertainty on LVLM's output. VAUQ operationalizes this insight through two key components.
First, we propose the Image-Information Score (IS), which quantifies the reduction in predictive uncertainty attributable to the image by comparing model behavior with and without visual input.
Moreover, to ensure that IS reflects semantically meaningful visual information rather than spurious background correlations, we introduce an unsupervised core-region masking strategy. By selectively masking the salient regions, VAUQ penalizes predictions that remain confident even after core visual evidence is removed.

We evaluate VAUQ across diverse benchmark datasets and widely used LVLMs, including LLaVA~\cite{liu2023visual}, Qwen2.5-VL~\cite{bai2025qwen2}, and InternVL3.5~\cite{wang2025internvl3}.
Across all settings, VAUQ consistently outperforms existing LLM- and LVLM-based self-evaluation methods. In the challenging counterfactual scenarios where visual grounding is essential, our approach
achieves a significant \textbf{+13.3\%} improvement in self-evaluation AUROC compared to state-of-the-art methods. We conduct comprehensive ablation studies that systematically analyze each component of VAUQ, demonstrating the necessity and robustness of each design choice.
Our key contributions are summarized as follows:

\begin{enumerate}
    \item We propose {VAUQ}, a novel vision-aware uncertainty quantification framework that enables LVLMs to perform reliable self-evaluation without relying on external models.
    \item We introduce an information-theoretic score together with a {core-region masking strategy} to effectively capture visual utilization in a label-free and training-free manner.
    \item We conduct extensive experiments across multiple LVLMs and benchmark datasets, achieving state-of-the-art performance and providing a rigorous analysis of the contributions of the proposed components.
\end{enumerate}

\section{Related Works}
\vspace{-1mm}

\paragraph{LLM self-evaluation} aims to assess the correctness of a model’s own outputs using only its internal knowledge, without relying on external supervision.
This is closely related to uncertainty quantification and hallucination detection, which can be broadly categorized as follows:
(1) {logit-based methods}, which use token-level probabilities as uncertainty scores~\cite{malinin2020uncertainty, orgad2024llms};
(2) {prompting-based methods}, which instruct models to explicitly express their uncertainty in natural language~\cite{kadavath2022language, lin2022teaching, lingenerating};
(3) {consistency-based methods}, which evaluate uncertainty by measuring the agreement across multiple generated responses~\cite{manakul2023selfcheckgpt, kuhnsemantic, chen2024inside, li2025semantic}; and
(4) {internal-state-based methods}, which leverage latent representations to assess hallucination~\cite{ren2022out, azaria2023internal,burns2022discovering, du2024haloscope, park2025steer, wang2024latent}.

\vspace{-1mm}

\paragraph{LVLM self-evaluation} \emph{remains largely underexplored}, as it is further complicated by the uncertainty introduced through the integration of the image information. Prior studies have examined token-level probabilities~\cite{zhou2023analyzing}, visual attention weights~\cite{jiang2024devils}, or latent representation~\cite{phukan2024beyond, yang2025nullu, park2025glsim, duan2025truthprint} to detect hallucinations at the object level.
However, these methods are limited to object tokens and cannot effectively assess response-level hallucination, which is more generally applicable across diverse vision-language tasks.
A few recent works attempt to detect response-level hallucination through linear probing~\cite{li-etal-2024-reference} or semantic-invariant perturbation~\cite{khan2024consistency, zhang2024vl}, but they rely on large labeled datasets, require computationally expensive multiple sampling, or external natural language inference modules.

In this paper, we propose an efficient framework for LVLM self-evaluation that identifies hallucinations based on the Image-information Score, enabling scalable and generalizable response-level assessment in a \textit{training-free} manner, \textit{without relying on external supervision}.

\section{Problem Setup}

\paragraph{Notation.}
Given an input image, the vision encoder processes it into a set of patch-level visual tokens.
These tokens are then projected into the language model’s embedding space through the multi-modal fusion module, resulting in a sequence of $N$ visual embeddings: 
$\mathbf{v} = \{v_1, \dots, v_N\} \in \mathbb{R}^{N \times d}$, where each $v_i$ corresponds to a transformed visual token of dimension $d$.
On the language side, the input text prompt  is tokenized and embedded into a sequence of language embeddings:
$\mathbf{t} = \{t_1, \dots, t_L\}\in \mathbb{R}^{L \times d}$, where $L$ is the prompt length.
The projected visual tokens $\mathbf{v}$ and the textual embeddings $\mathbf{t}$ are concatenated and passed as the input sequence to the language model.
The language model then generates a sequence of output tokens: 
$\mathbf{y}= \{y_1, \dots, y_M\}$, where each $y_i\in \mathcal{V}$ is drawn from a vocabulary space and $M$ is the output length.

In real-world deployment, LVLMs are often required not only to generate answers, but also to assess the reliability of answers in the absence of external supervision.  Self-evaluation is thus crucial for selective prediction, hallucination detection, and downstream decision-making under uncertainty. We provide the formal definition below.

\begin{definition}[\textbf{LVLM Self-Evaluator}]
\emph{
Let $\mathbf{x} = (\mathbf{v}, \mathbf{t})$ denote the multimodal input to an LVLM.
The goal of self-evaluation is to design a scoring function $s:\mathcal{X}\times\mathcal{Y}\rightarrow[0,1]$,
where $s(\mathbf{x}, \mathbf{y})$ measures the likelihood that the generated response $\mathbf{y}$ is hallucinated for the input $\mathbf{x}$.
Based on this score, we define the binary self-evaluator:
\begin{equation}
\label{equ:def}
G(\mathbf{x}, \mathbf{y}) =
\begin{cases} 
1, & \text{if } s(\mathbf{x}, \mathbf{y}) \geq \tau \\ 
0, & \text{otherwise},
\end{cases} 
\end{equation}
where $\tau \in [0,1]$ is a decision threshold.
A value of $G(\mathbf{x}, \mathbf{y}) = 1$ indicates that the response is classified as hallucinated (\ie, incorrect), while $G(\mathbf{x}, \mathbf{y}) = 0$ denotes a correct response.
}
\end{definition}

\section{Our Approach}

In this section, we introduce Vision-Aware Uncertainty Quantification (\textbf{VAUQ}), a training-free self-evaluation framework that explicitly measures an LVLM’s reliance on visual evidence when generating responses.
We first present our motivation in \cref{sec:motivation}, followed by a detailed walkthrough of the proposed method in \cref{sec:method}.

\begin{figure*}[t]
    \centering
    \includegraphics[width=\linewidth]{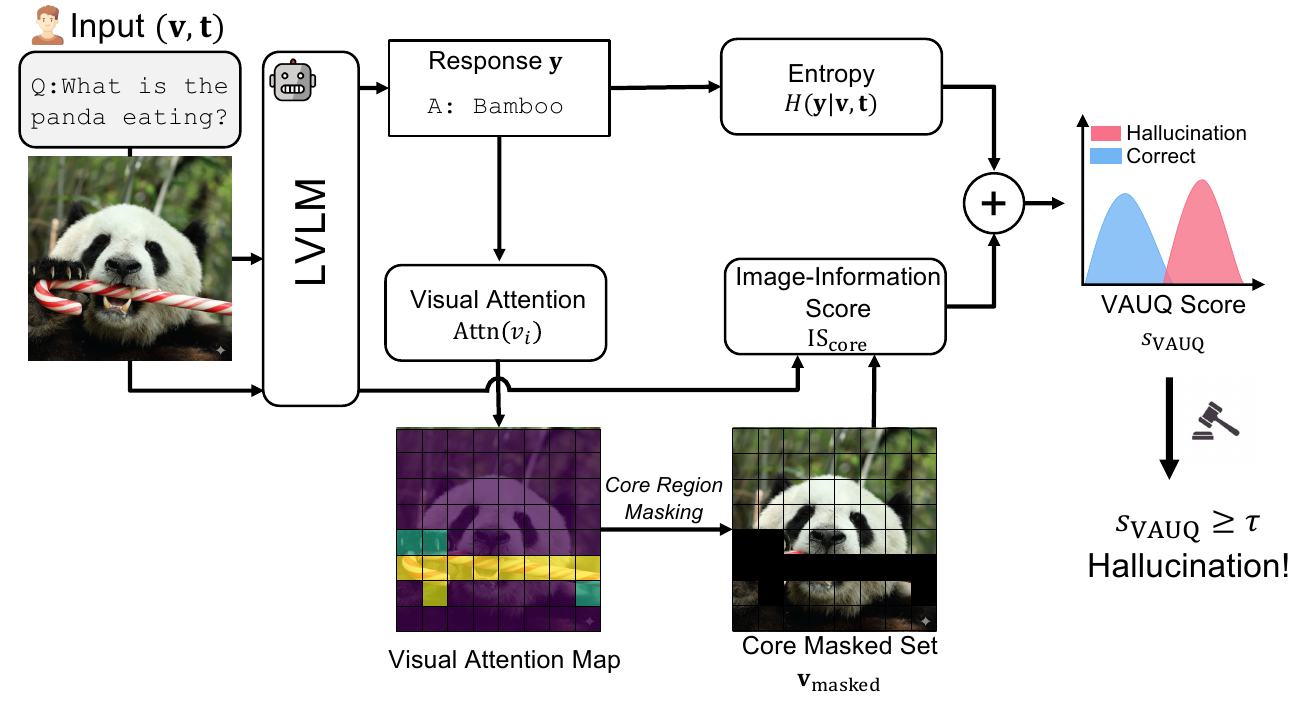}
   \caption{
    \textbf{Overall VAUQ Framework.}
    Given an input image-text pair $(\mathbf{v}, \mathbf{t})$, the LVLM generates a response $\mathbf{y}$.
    Based on the attention map  $\mathrm{Attn}(v_i)$, we perform unsupervised core region masking by covering the top-$K\%$ image patches, resulting in a core-masked set $\mathbf{v}_{\text{masked}}$.
    Using this masked input, we compute the core-masked Image-Information Score $\mathrm{IS}_{\text{core}}$.
    Finally, predictive entropy $H(\mathbf{y}\mid \mathbf{v}, \mathbf{t})$  and $\mathrm{IS}_{\text{core}}$ are combined to produce the VAUQ score $s_{\mathrm{VAUQ}}$ for self-evaluation.
    }
    \label{fig:3}
\end{figure*}
\subsection{Motivation: LLM-based Self-evaluation Methods Suffer from Language Prior}
\label{sec:motivation}

Recent works on uncertainty quantification or hallucination detection methods for LLMs demonstrate that models can self-evaluate the correctness of their responses through internal signals~\cite{lin2022teaching, wang2024latent}.
However, these methods are developed for text-only settings, and it remains unclear whether they can capture uncertainty that arises specifically from incorporating image information in vision-language tasks.
To investigate the effectiveness of LLM-based self-evaluation metrics in vision-language multimodal setups, we conduct a pilot study (Figure~\ref{fig:2}) on the ViLP dataset~\cite{luo2024probing}, which contains pairs of factual images aligned with common knowledge and counterfactual images paired with identical prompts.
Dataset details are provided in \cref{sec:dataset}.

We consider four representative LLM-based self-evaluation approaches: (1) {Length-normalized Entropy (Ent)}~\cite{malinin2020uncertainty}, which measures the length-normalized entropy of the next-token distribution; (2) {Verbalized Confidence (Verb)}~\cite{lin2022teaching}, obtained by parsing the model’s self-reported numerical confidence; (3) {Semantic Entropy (Sem)}~\cite{kuhnsemantic}, defined as the entropy over semantic clusters formed from multiple sampled responses; and (4) {EigenScore (Eigen)}~\cite{chen2024inside}, which quantifies representation-level variability via dominant eigenvalues of the hidden-state covariance matrix. 
All the formulations are provided in~\cref{sec:baselines}.

We observe that all four metrics show substantial performance degradation on counterfactual images, \eg, Entropy decreases by \textbf{-40.9\%} and EigenScore by \textbf{-26.0\%}, suggesting that {these methods are strongly driven by language priors and fail to reliably incorporate visual evidence} when measuring uncertainty. In particular, low predictive uncertainty does not necessarily indicate correct visual grounding, as an LVLM may remain confident even when its response is weakly supported, or contradicted, by the given image.
These findings motivate the need for new metrics that \emph{explicitly capture the use of visual evidence in LVLM self-evaluation}.
Accordingly, our research question is: 

\begin{center} \textbf{\textit{How can we effectively quantify an LVLM’s reliance on visual evidence for self-evaluation?}}
\end{center}

\subsection{Vision-Aware Uncertainty Quantification}
\label{sec:method}

To address the research question, we propose a vision-aware uncertainty quantification framework that explicitly measures how much visual information contributes to a model’s prediction. Our approach builds on the idea that visual evidence should reduce predictive uncertainty when it is informative and correctly utilized.
Accordingly, we propose the \emph{Image-Information Score} (IS), capturing the degree to which the image influences the model’s predictive uncertainty.

\begin{definition}[\textbf{Image-Information Score (IS)}]
\emph{Let $H(\mathbf{y} \mid \mathbf{v}, \mathbf{t})$\footnote{ 
$H(\mathbf{y} \mid \mathbf{v}, \mathbf{t}) = -\frac{1}{M} \sum_{i=1}^{M} \sum_{y \in \mathcal{V}} p(y_i = y \mid \mathbf{y}_{<i},\mathbf{v}, \mathbf{t}) \log p(y_i = y \mid  \mathbf{y}_{<i}, \mathbf{v}, \mathbf{t})$
}
denote the length-normalized conditional entropy
of the model’s predictive distribution given visual features $\mathbf{v}$ and text $\mathbf{t}$, and let $H(\mathbf{y}  \mid \varnothing, \mathbf{t})$ denote the entropy when visual tokens are removed.
The Image-Information Score is defined as:
\begin{equation} \label{equ:base_is}
\mathrm{IS_\text{blank}}
= H(\mathbf{y} \mid \varnothing, \mathbf{t})
- H(\mathbf{y} \mid \mathbf{v}, \mathbf{t}),
\end{equation}
}
\end{definition}
\noindent where a larger IS indicates that visual information substantially reduces predictive uncertainty, reflecting stronger visual grounding during the models' answer generation. %

\paragraph{Unsupervised Core Region Masking.}
While $\text{IS}_{\text{blank}}$ serves as an effective proxy for measuring image utilization, one limitation is that it can be sensitive to spurious correlations in the input (\eg, background artifacts)~\cite{yang2023mitigating}, which introduce noise unrelated to the core visual evidence (See~\cref{tab:masking} for the details).

To mitigate this challenge, we introduce an unsupervised core-region masking strategy that focuses on the most task-relevant visual regions for score computation. Intuitively, if a model truly relies on the visual evidence needed to answer a query, removing that evidence should significantly increase predictive uncertainty.
Since ground-truth evidence annotations are unavailable at inference time, we estimate these regions using the model’s visual attention weights. 
Prior analyses indicate that middle to later transformer layers provide the most informative alignment between visual tokens and semantic reasoning~\cite{jiang2024devils, park2025glsim}.
We therefore aggregate attention from a contiguous range of layers, and empirically verify the optimality of this design choice in Section~\ref{sec:ablation}~(\cref{fig:vis_attn_layer}).

Specifically, we define the interaction between a generated token \( y_i \) and visual information by summing the attention weights assigned to image tokens within an attention head \( h \) and layer \( \ell \):
\begin{equation}
\mathrm{Attn}(v_i) \triangleq \sum_{l=l_\text{s}}^{l_\text{e}} \sum_{h=1}^{H} \sum_{j=1}^{M} A^{(\ell,h)}(y_j, v_i),
\end{equation}
where \( A^{(\ell,h)}(y_j, v_i) \) denotes the attention weight from generated token \( y_j \) to image token \( v_i \) at the \(h\)-th attention head of the \(\ell\)-th layer, \( M \) is the number of generated tokens, \( H \) is the total number of attention heads, and \( l_s \) and \( l_e \) indicate the start and end indices of the transformer layers used for aggregation, respectively.

We then select the top \(K\%\) of image patches with the highest attention scores as $\mathbf{v}_{\text{top}}$:
\begin{equation}
\label{equ:topk}
\small
\mathbf{v}_{\mathrm{top}} \triangleq \left\{ v_i \;\middle|\; i \in \operatorname{TopK\%}\!\left(\{\mathrm{Attn}(v_i)\}_{i=1}^{N}\right) \right\},
\end{equation}
and define the \textit{remaining} set of visual tokens, $\mathbf{v}_{\mathrm{masked}}$, which is then used as input for core-masked IS computation:
\begin{align}
\label{equ:core}
\mathrm{IS}_{\text{core}}&=H(\mathbf{y} \mid \mathbf{v}_{\text{masked}},\mathbf{t})-H(\mathbf{y}\mid\mathbf{v},\mathbf{t}),
\end{align}
where $\mathbf{v}_{\mathrm{masked}} \triangleq \left\{ v_i \;\middle|\; v_i \notin \mathbf{v}_{\mathrm{top}} \right\}$. By masking the most attended regions, we intentionally remove the visual evidence the model is most likely to rely on, enabling a more precise assessment of whether its predictions are genuinely grounded.

\paragraph{Vision-Aware Uncertainty Quantification.}

Finally, we define the vision-aware uncertainty quantification score, $s_{\mathrm{VAUQ}}$, as a linear combination of the predictive entropy and the IS score with the core region masking:
\begin{equation}
\small
\label{equ:vauq}
\begin{aligned}
s_{\mathrm{VAUQ}}&(\mathbf{x}, \mathbf{y})
= H(\mathbf{y} \mid \mathbf{v}, \mathbf{t})
   - \alpha \cdot \mathrm{IS}_{\text{core}} \\
&=
\underbrace{(1+\alpha)\, H(\mathbf{y} \mid \mathbf{v}, \mathbf{t})}_{\text{predictive uncertainty}}
\;-\;
\underbrace{\alpha\, H(\mathbf{y} \mid \mathbf{v}_{\text{masked}}, \mathbf{t})}_{\substack{\text{penalize weak use of}\\\text{core visual information}}},
\end{aligned}
\end{equation}
where $\alpha$ is a weighting hyperparameter. This formulation can be interpreted as measuring predictive uncertainty while explicitly discounting confidence that is not supported by core visual evidence. 
When a model relies on visual information, masking the core regions leads to a significant increase in entropy, resulting in a lower $s_{\mathrm{VAUQ}}$ score and signaling a more reliable, well-grounded prediction. 
In contrast, when a model’s prediction is dominated by language priors, the IS score remains low—reflecting limited uncertainty reduction from visual input—which results in a relatively higher $s_{\mathrm{VAUQ}}$ value and indicates an increased risk of hallucination.

\begin{table*}[t!]
\centering
\small
\begin{tabular}{lcccc|cccc}
\toprule
& \multicolumn{4}{c}{\textbf{LLaVA-1.5-7B}} 
& \multicolumn{4}{c}{\textbf{LLaVA-1.5-13B}} \\
\cline{2-9}
\textbf{Method} 
& \textbf{ViLP} & \textbf{MMVet} & \textbf{VisualCoT} & \textbf{CVBench} 
& \textbf{ViLP} & \textbf{MMVet} & \textbf{VisualCoT} & \textbf{CVBench}  \\
\midrule

Perplexity            &  54.6 & 79.3  & 56.2 & 60.3  & 54.2  & 81.3  & 63.7  & 64.6  \\
Verbalized             &  56.3 & 71.7  & 49.9 & 57.6 & 52.4  & 67.2  & 54.4  & 55.4  \\
SVAR                  &  50.6 & 29.3  & 44.5 & 48.7   &  48.3 & 34.2   & 44.5 & 49.2   \\
Contextual Lens       &  56.7 & 70.8  & 58.3 & 51.1  & 60.4  & 61.4  & 52.5   &  58.6  \\
Chain-of-Embeddings   &  52.0 & 60.8  &  55.7    & 50.6  & 60.5  & 44.4   &  46.5  & 53.3  \\
EigenScore            &  63.2 & 78.2  & 74.7 & 65.8  &  61.1  & 85.1  & 75.5  &  58.1 \\
Semantic Entropy      &  \underline{63.7} & 81.3  & \underline{75.1} & 70.2  &  \underline{63.5} & \underline{86.4}  & 75.7  &  66.2 \\
VL-Uncertainty        &  55.6 & \textbf{82.3}  & 65.2 & \underline{71.1}  &  58.6 &  85.9 &  \underline{77.7} &  \underline{67.9} \\
\rowcolor{gray!10} \textbf{$\text{VAUQ}$ (Ours)} & \textbf{77.0}  & \underline{81.5} & \textbf{77.8}  & \textbf{73.2}  & \textbf{69.5}  & \textbf{88.6}  & \textbf{80.2}  & \textbf{68.3}  \\
\hline
\end{tabular}
\caption{\textbf{Main results with LLaVA.} Comparison with competitive self-evaluation methods across datasets. All values are AUROC (\%). Best results are in \textbf{bold}, and second-best results are \underline{underlined}.} \label{tab:main_results}
\end{table*}

\begin{table*}[t!]
\centering
\small
\begin{tabular}{lcccc|cccc}
\toprule
& \multicolumn{4}{c}{\textbf{Qwen-2.5-VL-7B}} 
& \multicolumn{4}{c}{\textbf{InternVL3.5-8B}} \\
\cline{2-9}
\textbf{Method} 
& \textbf{ViLP} & \textbf{MMVet} & \textbf{VisualCoT} & \textbf{CVBench} 
& \textbf{ViLP} & \textbf{MMVet} & \textbf{VisualCoT} & \textbf{CVBench}  \\
\midrule

Perplexity            & 55.0  & \underline{76.6}   & 56.0  & 64.8  & 55.3  & 67.5  & 62.3  & 66.4   \\
Verbalized             &  55.3 & 51.9  & 54.7  & 56.3  & 48.7  & 59.4  & 54.7  & 60.1   \\
SVAR          &  49.6 & 54.6  & 46.7  & 61.6  & 51.6 & 49.7   & 56.7   & 50.5   \\
Contextual Lens & \underline{61.3}  & 65.5  &  \underline{65.0} & 54.4  & 51.3  & 59.6   & 55.1   & 48.2   \\
Chain-of-Embeddings & 47.4  & 59.8 & 44.3  & 49.7  & 50.1  & 52.8   &  61.4  & 53.6   \\
EigenScore            &  53.0 & 60.8  &  51.1 & 50.9  & 59.3   & 64.1  & \underline{74.5}   &  61.0 \\
Semantic Entropy      & 52.0  & 60.1  &  53.3 & 50.9  & 64.2  &  70.4 & 66.3  & \underline{73.7}   \\
VL-Uncertainty      & 57.9  & 69.7  &  62.3 & \underline{69.7}  &  \textbf{67.4} & \underline{75.7}  &  65.8 & 72.0  \\
\rowcolor{gray!10} \textbf{$\text{VAUQ}$ (Ours)} & \textbf{64.1}  & \textbf{78.3}  & \textbf{68.0}  &  \textbf{69.8}  & \underline{65.2} &  \textbf{75.8} &  \textbf{77.2} & \textbf{74.7}  \\
\hline
\end{tabular}
\caption{\textbf{Main results using QwenVL-2.5-7B and InternVL-3.5-8B.} All values are AUROC (\%). Best results are in \textbf{bold}, and second-best results are \underline{underlined}.} \label{tab:main_results2}
\end{table*}

\section{Experiments}
\subsection{Experimental Setup}
\paragraph{Datasets and Models.}
We evaluate our method on three free-form visual question answering datasets: ViLP~\cite{luo2024probing}, MMVet~\cite{yu2023mm}, and VisualCoT~\cite{shao2024visual}, and one multiple-choice benchmark: CVBench~\cite{tong2024cambrian}.
These datasets broadly cover the core challenges faced by deployed LVLMs, spanning language-prior dominance, multi-capability reasoning, evidence localization, and vision-centric perceptual reasoning.

We conduct experiments on three representative LVLMs: LLaVA-1.5-\{7B, 13B\}~\cite{liu2023visual}, Qwen-2.5-VL-7B~\cite{bai2025qwen2}, and InternVL3.5-8B~\cite{wang2025internvl3}. 
 Implementation details are provided in~\cref{sec:implementation}.

\paragraph{Baselines.}
We compare our method against eight representative self-evaluation baselines spanning both LLM- and LVLM-based approaches. 
The LLM-based methods include Perplexity~\cite{ren2022out}, Verbalized Confidence~\cite{kadavath2022language}, Chain-of-Embeddings~\cite{wang2024latent}, EigenScore~\cite{chen2024inside}, and Semantic Entropy~\cite{kuhnsemantic}. 
LVLM-based methods include SVAR~\cite{jiang2024devils}, Contextual Lens~\cite{phukan2024beyond}, and VL-Uncertainty~\cite{zhang2024vl}.

\paragraph{Evaluation Metrics.}
Following prior work~\cite{du2024haloscope, park2025steer}, we evaluate performance using the area under the receiver operating characteristic curve (AUROC). 
Ground-truth correctness labels for model responses are annotated using GPT-5~\cite{openai2025gpt5}.

\subsection{Main Experiments}

\paragraph{VAUQ achieves state-of-the-art performance.} In~\cref{tab:main_results,tab:main_results2}, we compare VAUQ with competitive self-evaluation approaches from prior work.
VAUQ achieves state-of-the-art performance, substantially outperforming existing methods on LLaVA, Qwen, and InternVL models.
We observe that both LLM-based scoring approaches and object-level hallucination detectors (\eg, SVAR and Contextual Lens) often exhibit inconsistent performance across architectures and data distributions.
In contrast, VAUQ more consistently delivers strong results across model families, scales, and four benchmark datasets.

Notably, VAUQ improves over a representative LLM-based method Semantic Entropy by \textbf{+13.4\%} on ViLP with LLaVA-1.5-7B.
This highlights the importance of explicitly modeling visual evidence for uncertainty quantification for LVLMs.
By jointly capturing predictive uncertainty and image-information utilization, VAUQ offers a more reliable and interpretable self-evaluation signal than prior text-centric methods.
Moreover, VAUQ outperforms the previous state-of-the-art LVLM self-evaluation method, VL-Uncertainty, by \textbf{+21.4\%} on ViLP and \textbf{+12.6\%} on VisualCoT.
While VL-Uncertainty estimates uncertainty by measuring semantic consistency across multiple sampled outputs and relies on external modules, VAUQ instead leverages internal model signals, requiring neither multiple sampling nor external components.

\subsection{Ablation Studies}
\label{sec:ablation}

\paragraph{How effective is core-region masking?}
\begin{table}[t!]
\centering
\resizebox{\columnwidth}{!}{
\begin{tabular}{l|ccc}
\textbf{Method}      & \textbf{LLaVA-1.5}  & \textbf{QwenVL-2.5} & \textbf{InternVL3.5} \\ \hline
w. $\text{IS}_{\text{blank}}$       &   75.2   & 66.9   &     75.9    \\
w. $\text{IS}_{\text{rand}}$       &  73.3    &    65.1    & 74.0        \\
w. $\text{IS}_{\text{GT}}$  &   78.6   &   69.2    &  77.7        \\ 
\rowcolor{gray!10} w. \textbf{$\text{IS}_{\text{core}}$}        & 77.8             &    68.0   & 77.2 \\
\hline
\end{tabular}}
\caption{AUROC (\%) of VAUQ score under different masking strategies on the VisualCoT dataset.} \label{tab:masking}
\end{table}

To investigate how masking strategies affect performance, we use the VisualCoT dataset~\cite{shao2024visual}, which provides ground-truth bounding boxes for the visual evidence required to answer each query.
We consider three variants of masking: (1) \textit{Blank}, which replaces the entire image with a blank input; (2) \textit{Random}, which applies a random mask to the image; and (3) \textit{GT (Oracle)}, which masks out the pixels inside the ground-truth evidence region.

The results in~\cref{tab:masking} show that masking the ground-truth evidence region ($\text{IS}_{\text{GT}}$) yields higher AUROC than the blank-image baseline ($\text{IS}_{\text{blank}}$), indicating that fine-grained and semantically relevant visual content is crucial for VAUQ score computation.
In contrast, random masking ($\text{IS}_{\text{rand}}$) leads to degraded performance relative to the blank baseline, suggesting that indiscriminate perturbations disrupt the model's ability to extract meaningful visual cues.
These findings highlight the importance of carefully selecting which visual regions to mask when estimating image-information utilization.
Our method ($\text{IS}_{\text{core}}$) achieves performance comparable to the GT (Oracle) setting, demonstrating that our approximated masking region effectively captures the core visual evidence.

\paragraph{Can visual attention weight capture the correct region?}

\begin{figure}[t!]
    \centering
\includegraphics[width=\linewidth]{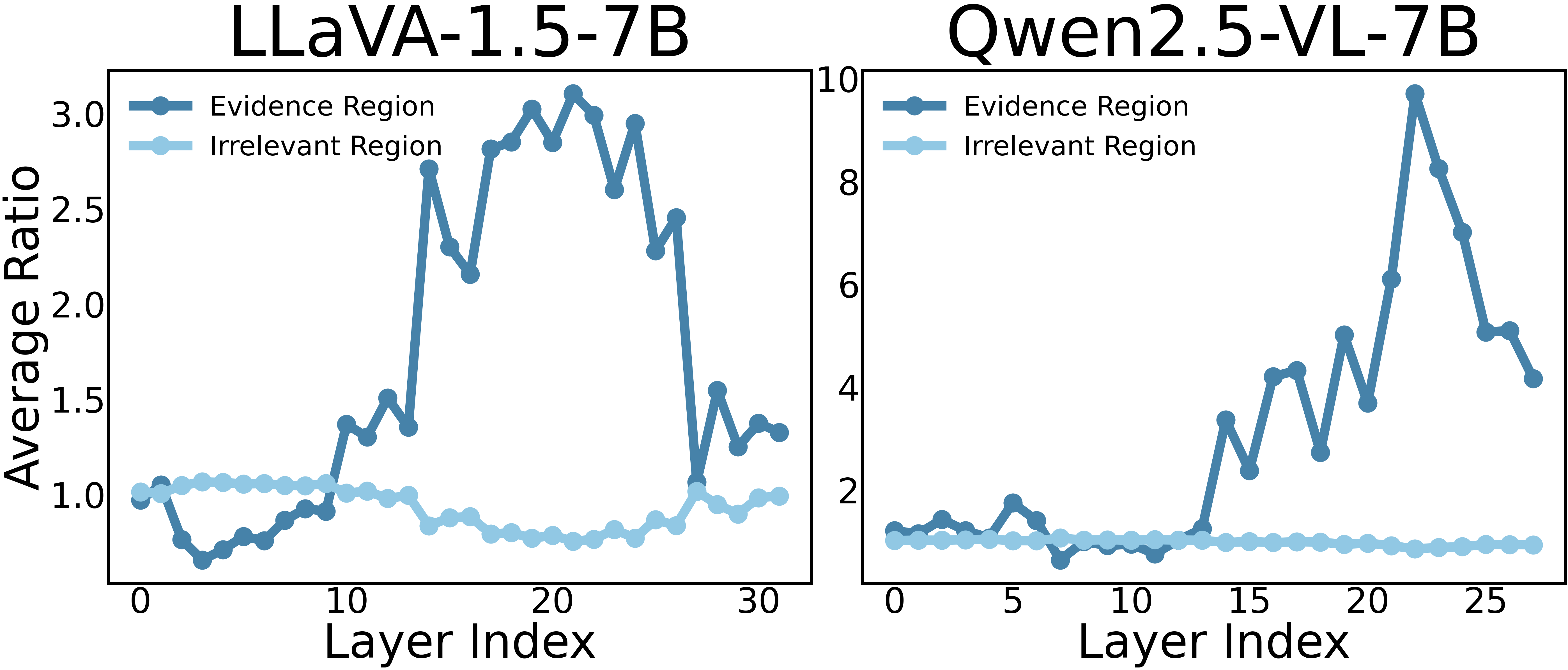}
    \caption{Visual attention ratios over evidence and irrelevant regions on the VisualCoT dataset.}
    \label{fig:vis_attn_layer}
\end{figure}

We investigate whether visual attention weights can capture the correct evidence region in the input image. 
Using the ground-truth evidence regions provided in the VisualCoT dataset, we compute the summed visual attention weight within the evidence region and within the irrelevant region, and then normalize each by the number of patches in the corresponding region. 
In~\cref{fig:vis_attn_layer}, we visualize the normalized attention weights for LLaVA-1.5-7B and Qwen2.5-VL-7B on VisualCoT.
For both models, the early layers (0--10) struggle to focus on the correct region, while middle to later layers capture the evidence region more reliably. 
This observation is consistent with prior findings that intermediate layers of LVLMs are primarily responsible for processing visual information, and it further supports our design choice to use intermediate layers in our approach.

\paragraph{Visualization results of core region masking.}

\begin{figure}[t!]
    \centering
\includegraphics[width=\linewidth]{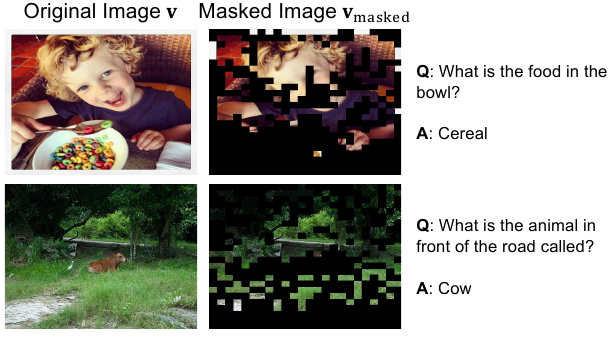}
    \caption{Qualitative examples of core region masking using LLaVA-1.5-7B.}
    \vspace{-0.2cm}
\label{fig:qual_core}
\end{figure}

We visualize masked images produced by the core region masking strategy using visual attention aggregated from the 10th-25th layers of LLaVA-1.5-7B, with $K=60$.
As shown in~\cref{fig:qual_core}, the masking strategy defined in~\cref{equ:topk} effectively masks out semantically important regions required for reasoning, such as the cereal or the cow in the original image.
These examples demonstrate that leveraging intermediate-layer visual attention can reliably identify meaningful visual regions.
Additional qualitative results are provided in~\cref{sec:qual_abl}.
\

\paragraph{Complementary roles of entropy and image information score.}
In~\cref{tab:is_comparison}, we conduct a component analysis of VAUQ on the ViLP dataset using LLaVA-1.5-7B and Qwen2.5-VL-7B.
Entropy performs well on the factual split but degrades substantially on the counterfactual split due to its reliance on language priors.
In contrast, $\mathrm{IS}_{\text{core}}$ achieves moderate performance on factual samples while exhibiting strong performance on the counterfactual split, where visual evidence is essential.
By combining these two signals, VAUQ achieves robust and balanced performance across both factual and counterfactual settings.
These results demonstrate that predictive entropy and image-information utilization capture complementary aspects of uncertainty, making VAUQ well-suited for mixed real-world distributions.

\begin{table}[t!]
\centering
\resizebox{\columnwidth}{!}{
\begin{tabular}{l|cc|cc}

& \multicolumn{2}{c|}{\textbf{Factual}} & \multicolumn{2}{c}{\textbf{Counterfactual}} \\ 
\cline{2-5}
\textbf{Method} & \textbf{LLaVA} & \textbf{QwenVL} & \textbf{LLaVA} & \textbf{QwenVL} \\ 
\hline
Entropy      & 83.2 & 68.4 & 48.4 & 44.5 \\
$\text{IS}_{\text{core}}$  & 60.9 & 56.5 & 75.2& 63.0\\
$\text{Entropy} - \alpha \text{IS}_{\text{core}}$  & 83.6 & 68.1 & 70.4& 60.1  \\
\hline
\end{tabular}
}
\caption{Component analysis of VAUQ on the ViLP dataset across factual and counterfactual splits.} \label{tab:is_comparison}
\end{table}

\begin{figure*}[t!]
    \centering

    \begin{subfigure}{0.32\textwidth}
        \centering
        \includegraphics[width=\linewidth]{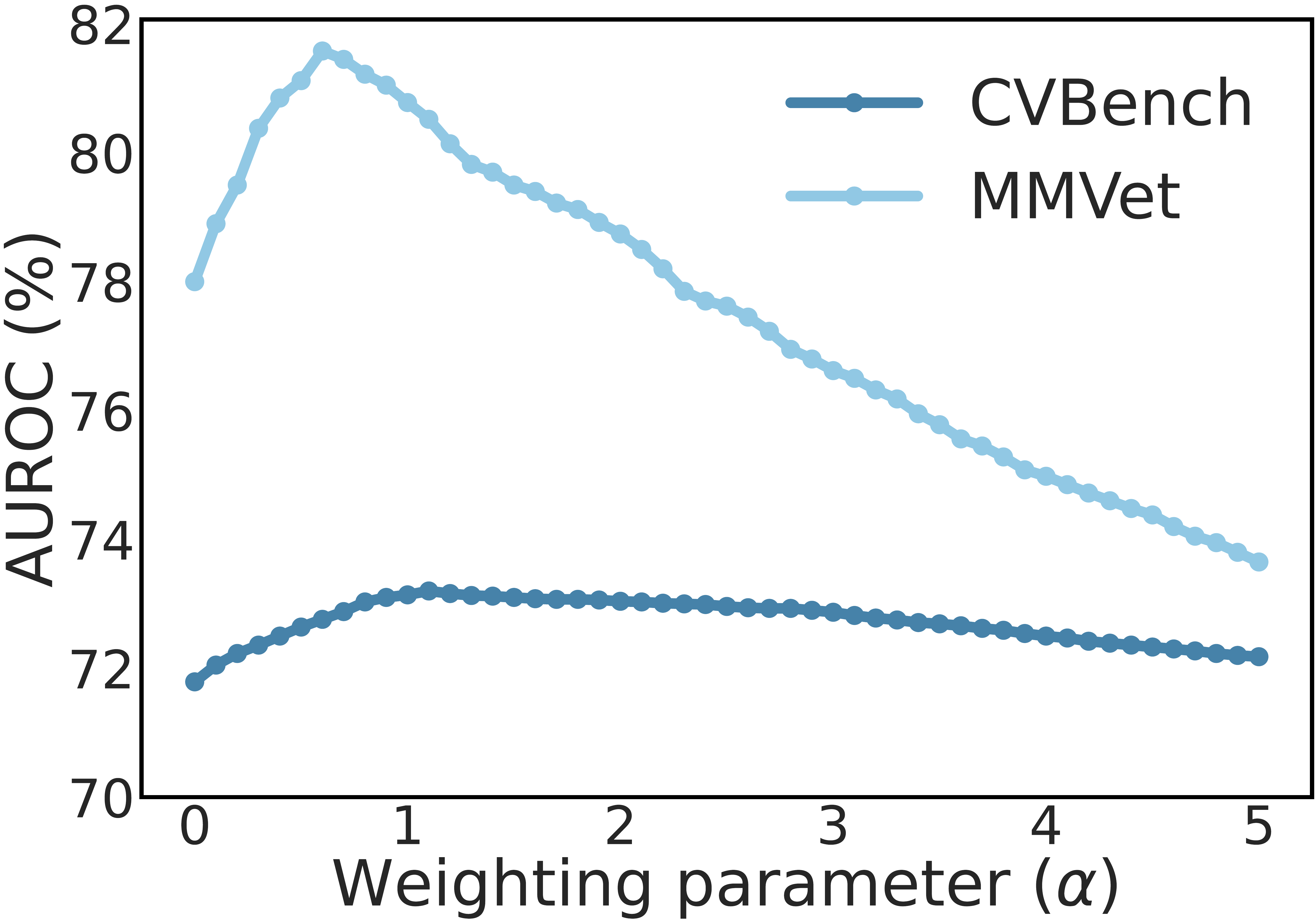}
        \caption{Weighting parameter $\alpha$}
        \label{fig:alpha}
    \end{subfigure}
    \hfill
    \begin{subfigure}{0.32\textwidth}
        \centering
        \includegraphics[width=\linewidth]{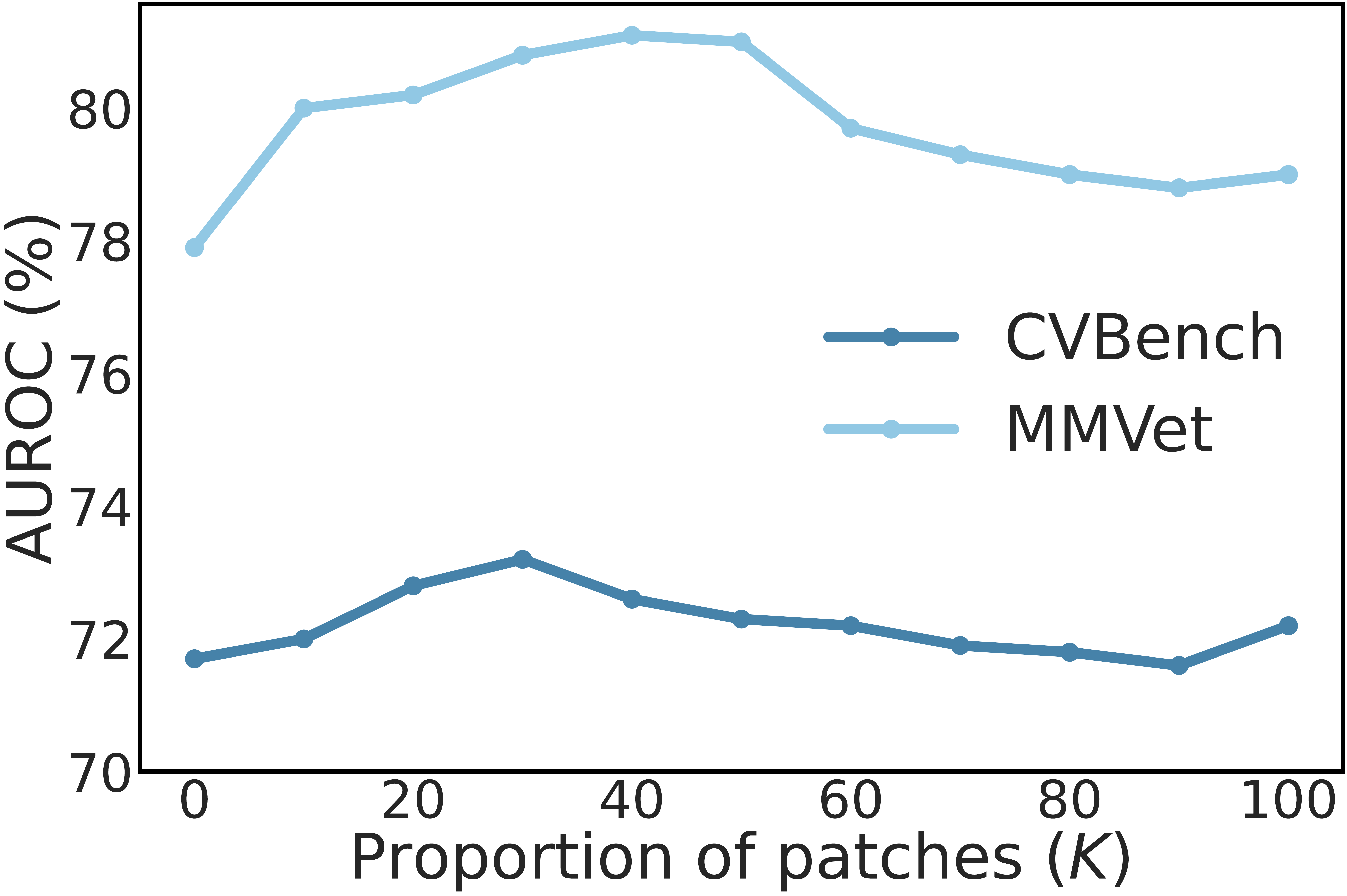}
        \caption{Proportion of patches $K$}
        \label{fig:k}
    \end{subfigure}
    \hfill
    \begin{subfigure}{0.32\textwidth}
        \centering
        \includegraphics[width=\linewidth]{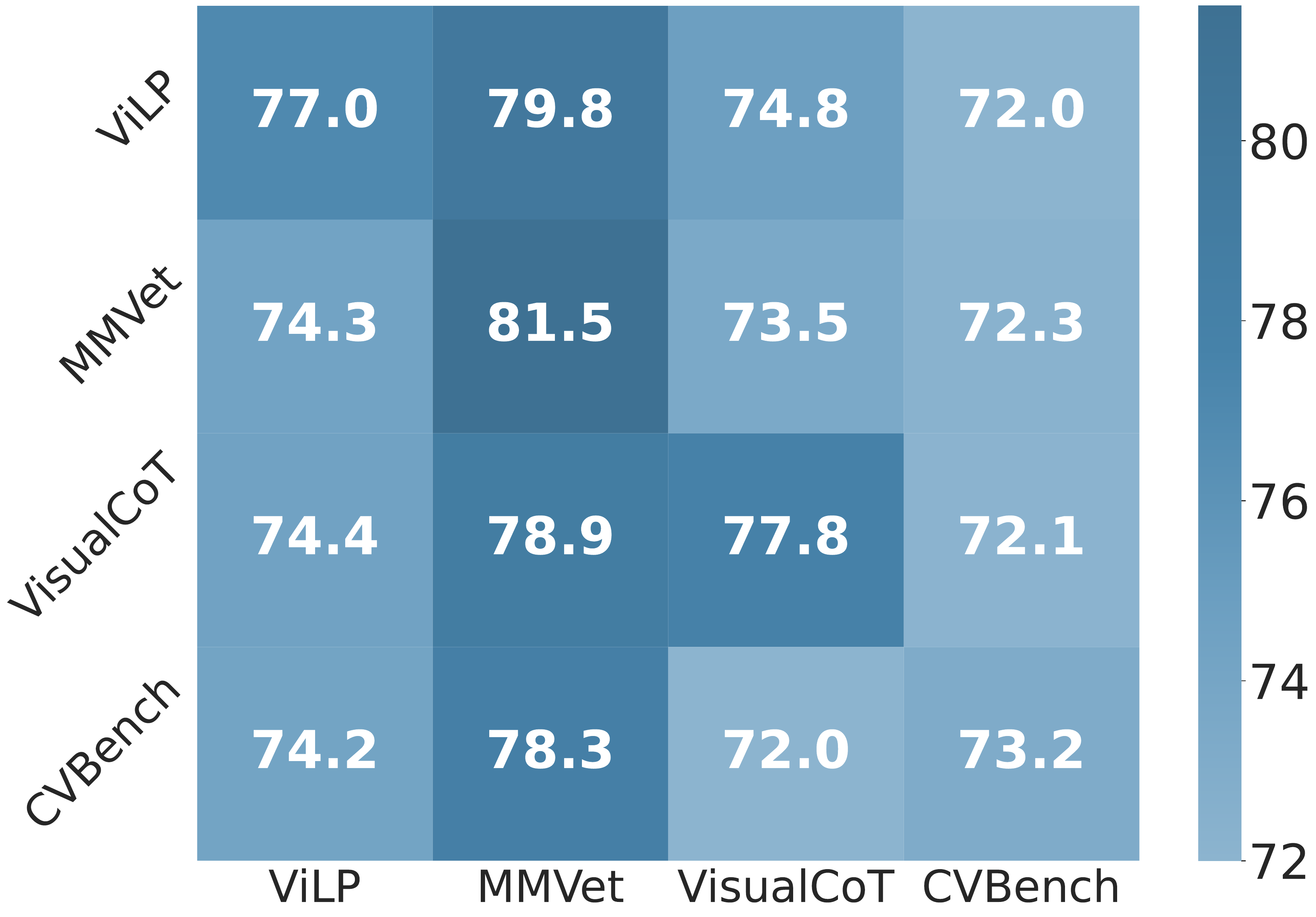}
        \caption{Generalization results}
        \label{fig:generalization}
    \end{subfigure}

\caption{(a) Effect of the weighting parameter $\alpha$ in~\cref{equ:vauq}; (b) effect of the proportion of masked image patches $K$ in~\cref{equ:topk}; (c) generalization performance across datasets.}
\vspace{-0.5cm}
    \label{fig:3fig}
\end{figure*}

\paragraph{Inference efficiency of VAUQ.}

\begin{table}[t!]
\centering
\resizebox{\columnwidth}{!}{
\begin{tabular}{l|cc|cc}

& \multicolumn{2}{c|}{\textbf{LLaVA-1.5-7B}} & \multicolumn{2}{c}{\textbf{Qwen2.5-VL-7B}} \\ 
\cline{2-5}
\textbf{Method} & \textbf{Time (s)} & \textbf{AUC (\%)} & \textbf{Time (s)} & \textbf{AUC (\%)} \\ 
\hline
SVAR     & 0.39 & 50.6 & 1.59 & 49.6 \\
Verbalized  & 0.58 & 56.3 & 1.82 & 55.3 \\
EigenScore    & 5.86 & 63.2 & 8.77& 53.0 \\ 
Semantic Entropy     & 7.05& 63.7 & 12.4& 52.0 \\ 
VL-Uncertainty   & 13.6 &  55.6    &20.2  & 57.9 \\
 \rowcolor{gray!10} \textbf{Ours}   & 0.73 &   77.0   & 2.16 & 64.1 \\
\hline
\end{tabular}
}
\caption{Average per-sample inference time on the ViLP dataset.}
\label{tab:inference}
\end{table}

In~\cref{tab:inference}, we evaluate the inference efficiency of VAUQ on the ViLP dataset.
Generating a response of length $M$ requires $M$ forward passes under standard autoregressive decoding.
VAUQ introduces only a constant number of additional forward passes for uncertainty estimation (e.g., computing scores with masked visual inputs), without requiring any additional autoregressive generation.
Consequently, the overall inference complexity of VAUQ remains linear in the output length, $O(M)$, with a small constant overhead.
In contrast, multi-sampling--based approaches require generating $A$ independent responses, resulting in $O(A \cdot M)$ forward passes, where $A$ is the number of samples and often exceeds five in practice~\cite{zhang2024vl}.
Consequently, VAUQ achieves a 94.6\% reduction in per-sample inference time compared to VL-Uncertainty, while delivering a +21.4\% AUROC improvement in self-evaluation performance on the LLaVA-1.5-7B model. 
Overall, VAUQ maintains comparable per-sample inference time while providing stronger self-evaluation accuracy, demonstrating a favorable efficiency-performance trade-off across models.

\paragraph{How does the weighting parameter $\alpha$ affect performance?}

In~\cref{fig:alpha}, we present the performance across different values of the weighting parameter $\alpha$. 
We search $\alpha$ over the range $[0, 5]$ with intervals of $0.1$. 
Larger values of $\alpha$ place greater emphasis on the IS score relative to predictive entropy. 
In practice, moderate values of $\alpha$ between 0.5 and 1.5 yield the best performance, suggesting that predictive entropy and IS provide complementary and balanced contributions. 
Tasks that rely more heavily on visual evidence tend to prefer larger $\alpha$ values (\eg, CVBench), while datasets requiring less visual grounding benefit from smaller values.

\paragraph{How does the proportion of the masked image patches $K$ affect performance?}
We examine how the proportion of selected image patches \(K\%\) used for masking influences model performance (see~\cref{fig:k}).
We search $K$ over the range [0, 100] with intervals of 10. 
We vary \(K\) from 0 to 100 in increments of 10. 
Overall, moderate values of $K$  yield stable performance across datasets, with optimal performance achieved at $K=30$ on CVBench and $K=40$ on MMVet.
In contrast, small masking regions tend to degrade performance, since they concentrate on overly specific areas or introduce noisy masks.

\paragraph{Parameter generalization across data distributions.}
While VAUQ demonstrates strong overall performance, we further investigate its ability to generalize across different data distributions.
As shown in~\cref{fig:generalization}, we evaluate the generalization capability of VAUQ using the LLaVA-1.5-7B model by transferring hyperparameters---including the weighting parameter~$\alpha$, the proportion of masked image patches~$K$---from a source in-distribution (ID) dataset to various target out-of-distribution (OOD) datasets, and computing their corresponding VAUQ scores.
The results show that VAUQ transfers robustly across diverse datasets. Notably, it achieves an AUROC of 72.3\% on CVBench even when the hyperparameters are selected from MMVet, closely matching the performance obtained when tuned directly on CVBench (73.2\%).
This transferability underscores VAUQ’s practical potential for real-world LVLM applications, enabling effective self-evaluation even under significant domain shifts.

\section{Conclusion}

We introduced \textbf{VAUQ}, a vision-aware uncertainty quantification framework for LVLM self-evaluation that explicitly accounts for how much a model relies on visual evidence when assessing the reliability of its own outputs.
Central to our approach is the Image-Information Score (IS), which measures the reduction in predictive uncertainty attributable to the image, and a core-region masking strategy that suppresses the influence of visually irrelevant regions during IS computation.
Extensive experiments across multiple LVLM architectures and benchmark datasets show that VAUQ consistently outperforms existing LLM- and LVLM-based self-evaluation methods. We hope our work will encourage future research on reliable and vision-aware self-evaluation in multimodal models.

\section*{Limitations}

Although VAUQ demonstrates robust performance across multiple LVLMs and benchmark datasets, it relies on a small set of global hyperparameters.
While we show that VAUQ is relatively stable under moderate variations of these values, the optimal configuration can still vary across datasets and even individual examples, depending on how strongly a task relies on visual versus linguistic information.
Hence, designing adaptive or sample-specific strategies for tuning these hyperparameters remains an interesting direction for future work.

\section*{Ethical Considerations}
VAUQ aims to support more reliable use of vision–language models by providing a lightweight, training-free self-evaluation signal.
Such a signal may be useful for identifying potentially unreliable outputs and for supporting selective prediction or human review in practical deployments.
At the same time, VAUQ is not a comprehensive safety mechanism and should not be treated as a definitive measure of correctness.
We view it as a complementary tool that is best used alongside existing safeguards and human oversight, particularly in sensitive or high-stakes settings.

\section*{Acknowledgements}

We gratefully acknowledge Lin Long, Yu Wang, Wendi Li, and Dahye Kim for their valuable comments on the draft, and the anonymous reviewers for their constructive feedback.
Seongheon Park, Changdae Oh, Hyeong Kyu Choi, and Sharon Li are supported in part by the AFOSR Young Investigator Program under award number FA9550-23-1-0184, National Science Foundation under awards IIS-2237037 and IIS-2331669, Office of Naval Research under grant number N00014-23-1-2643, Schmidt Sciences Foundation, Open
Philanthropy, Alfred P. Sloan Fellowship, and gifts from Google and Amazon.

\bibliography{custom}

\newpage

\appendix

\newpage
\section{Implementation Details}
\label{sec:implementation}

We implement our method using greedy decoding with a maximum generation length of 128 tokens.
For implementation efficiency, rather than modifying the raw image inputs corresponding to \(\mathbf{v}_{\text{masked}}\), we mask the attention weights associated with visual tokens in \(\mathbf{v}_{\mathrm{top}}\) when computing \(\mathrm{IS}_{\text{core}}\), following the attention knockout strategy~\cite{geva-etal-2023-dissecting,kaduri2025s}.
The weighting parameter $\alpha$ and the proportion of masked image patches $K$ used to compute the VAUQ score are selected based on a held-out validation set, as described in~\cref{sec:hyperparams}.
The layer index range $(l_s,l_e)$ is chosen heuristically based on empirical observations, as illustrated in \cref{fig:vis_attn_layer}.
For all experiments, we report results averaged over three random seeds.
All experiments are conducted using Python 3.11.11 and PyTorch 2.6.0 \cite{paszke2019pytorch} on a single NVIDIA A100 GPU with 80GB of memory.

\paragraph{Ground-truth labeling.}
For free-form visual question answering datasets—MMVet, VisualCoT, and ViLP—we employ GPT-5~\cite{openai2025gpt5} as an evaluator under the \emph{LLM-as-a-judge} paradigm~\cite{zheng2023judging} to annotate model outputs.
Specifically, we assess the correctness of LVLM-generated responses by determining their semantic equivalence to the corresponding gold-standard answers. We use the following evaluation prompt:

\begin{tcolorbox}[colback=gray!10, colframe=black, title=Input prompt for GPT-5-based labeling]
\textbf{Prompt:}\\
Ground truth: \textcolor{magenta}{\{ground\_truth\}}. \\
Model answer: \textcolor{magenta}{\{model\_answer\}}. \\
Please verify whether the model answer matches the ground truth. Respond with either \texttt{Correct} or \texttt{Wrong} only.
\end{tcolorbox}

A response is labeled as correct if the judge outputs \texttt{Correct}, and as a hallucination otherwise. To improve label consistency, we sample three independent judgments and assign the final label via majority voting.
For the multiple-choice dataset CVBench, we use exact answer matching for labeling.

\section{Qualitative Analysis}
\label{sec:qual_abl}

\subsection{Qualitative Example}
\begin{figure}[h!]
    \centering
    \includegraphics[width=\linewidth]{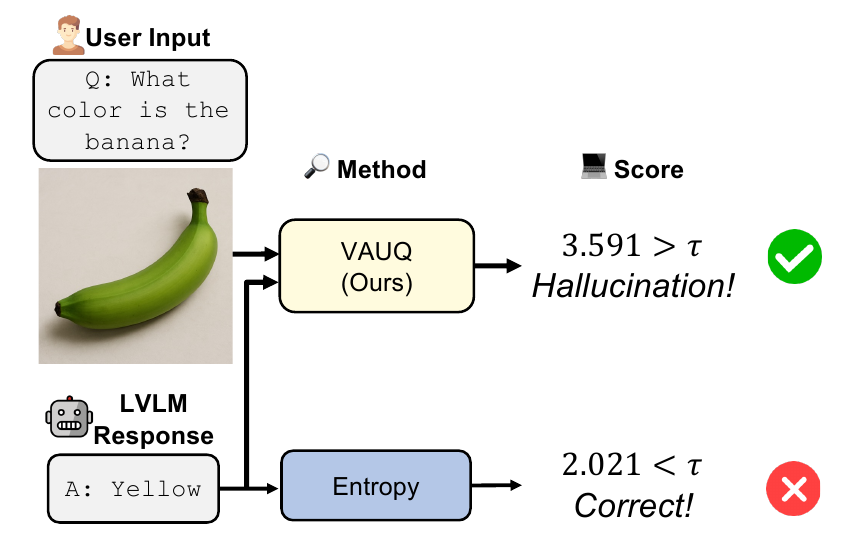}
    \caption{Qualitative example comparing our method with the entropy baseline.}
    \label{fig:fig_supp}
\end{figure}

In~\cref{fig:fig_supp}, we show a qualitative example comparing our method with the entropy baseline. The entropy baseline produces a lower uncertainty score than the threshold ($\tau$), predicting the generated response as correct, which illustrates the language prior problem inherent in LLM-based self-evaluation. In contrast, VAUQ explicitly accounts for visual information and produces a higher uncertainty score, accurately identifying the hallucinated output.

\subsection{Case Studies of Core Region Masking}

\begin{figure}[h!]
    \centering
\includegraphics[width=\linewidth]{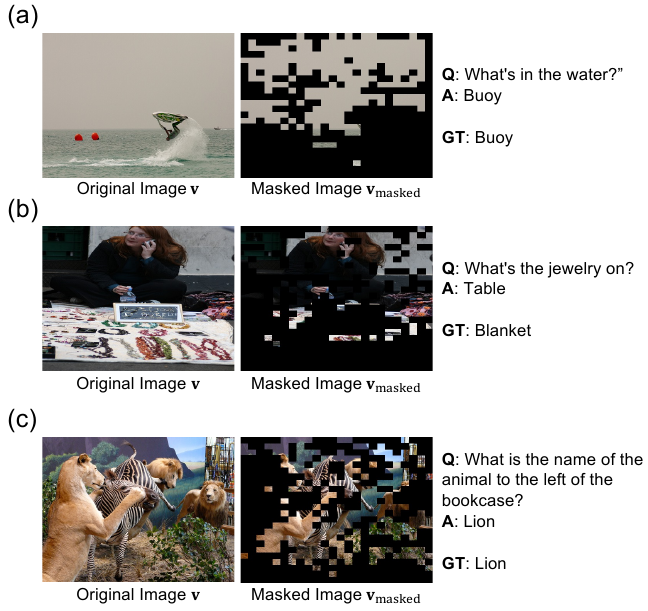}
    \caption{Qualitative case studies of core-region masking using LLaVA-1.5-7B, illustrating (a) correct response, (b) incorrect response, and (c) failure mode.}
    \label{fig:qual_appen}
\end{figure}

We conduct a qualitative analysis of core region masking under three scenarios: (a) correct response, (b) incorrect response, and (c) failure mode analysis, as shown in~\cref{fig:qual_appen}.
We visualize the masked images using visual attention from the 10th–25th layers of LLaVA-1.5-7B, with $K=60$.
For (a), when the response is correct, the model successfully identifies and masks out semantically meaningful regions.
For (b), although the model correctly captures meaningful regions of the image, it fails to reason about the output, indicating that accurate region identification alone is insufficient for correct reasoning. In this case, VAUQ captures information more effectively, as reflected by its performance compared to SVAR and our method in~\cref{tab:main_results,tab:main_results2}.
For (c), we observe a failure case in which multiple dominant objects are present in the image, causing visual attention to miss some relevant information.

\section{Dataset Details}
\label{sec:dataset}
\paragraph{ViLP~\cite{luo2024probing}} is a free-form vision--language evaluation dataset designed to probe language priors in LVLMs.
It contains 300 carefully constructed questions, each paired with three distinct image–answer sets: one factual answer that can be inferred from language context alone, and two counterfactual answers that require grounded visual reasoning to answer correctly, yielding 900 Question–Image–Answer (QIA) triplets in total.
In our work, we use the paired factual–counterfactual subset comprising 600 QIA triplets.

\paragraph{MMVet~\cite{yu2023mm}} is a vision--language evaluation benchmark designed to assess the integrated vision–language capabilities of LVLMs.
It defines six core vision–language capabilities—recognition, OCR, knowledge, spatial awareness, language generation, and math—and evaluates 16 capability integrations derived from their combinations.
MM-Vet consists of 218 open-ended questions paired with images, where each question requires one or more capabilities to answer correctly.

\paragraph{VisualCoT~\cite{shao2024visual}} is a free-form vision--language dataset designed to support and evaluate visual chain-of-thought reasoning in multimodal large language models.
Each example is annotated with a question, answer, and an intermediate bounding box that highlights the key evidence image region required for reasoning.
The dataset spans multiple domains, including text/document understanding, charts, general VQA, fine-grained recognition, and relational reasoning.
In our experiments, we sample 1.5K examples from the full VisualCoT dataset for evaluation.

\paragraph{CVBench~\cite{tong2024cambrian}} is a multiple-choice, vision-centric evaluation benchmark designed to assess fundamental visual understanding in LVLMs.
It repurposes classic vision benchmarks into vision--language questions that probe core 2D and 3D perception abilities, including spatial relationships, object counting, depth order, and relative distance.
CVBench consists of 2,638 manually inspected examples.

\section{Additional Analysis}
\subsection{Evaluation on HallusionBench}

To further validate our method, we evaluate on HallusionBench~\cite{guan2024hallusionbench} using LLaVA-1.5-7B and Qwen-2.5-VL-7B with the same baselines as in the main experiments. As shown in Table~\ref{tab:hallusion}, our method outperforms multi-sampling baselines, surpassing VL-Uncertainty by +1.9\% AUROC on LLaVA-1.5-7B and Semantic Entropy by +0.3\% on Qwen-2.5-VL-7B, and consistently outperforms all baselines across both models.

\begin{table}[h!]
\centering
\resizebox{0.8\columnwidth}{!}{
\begin{tabular}{l|c|c}

\textbf{Method} & \textbf{LLaVA} & \textbf{Qwen} \\ 
\cline{2-3}
\hline
Perplexity     & 56.4 & 69.4  \\
Verbalized     & 49.1 & 57.9  \\
SVAR     & 60.2 & 63.3 \\
Contextual Lens    & 54.9 & 58.3 \\
Chain-of-Embeddings    & 52.1  & 61.0 \\ 
Semantic Entropy     & 64.3 & 74.0  \\ 
VL-Uncertainty   &  65.1 & 73.5 \\
 \rowcolor{gray!10} \textbf{Ours}   & 67.0 & 74.3 \\
\hline
\end{tabular}
}
\caption{Experiment results on HallusionBench.}
\label{tab:hallusion}
\end{table}

\subsection{Comparison with Other Masking Strategies}

We compare our attention-based core-region masking against two alternative strategies: an embedding-based baseline (Contextual Lens) and a Grad-CAM--based~\cite{selvaraju2020grad} masking variant.

\paragraph{Localization Quality.}
To quantify localization quality, we measure the overlap between attention-derived core regions and ground-truth object masks on ImageNet-S~\cite{gao2022large}, which provides pixel-level annotations. As shown in Table~\ref{tab:localization}, our attention-based masking achieves consistently stronger alignment with ground-truth object regions than the embedding baseline across all metrics.

\begin{table}[h!]
\centering
\resizebox{0.95\columnwidth}{!}{
\begin{tabular}{l|c|c|c}
\textbf{Method} & \textbf{Pixel Acc.}$\uparrow$ & \textbf{mIoU}$\uparrow$ & \textbf{mAP}$\uparrow$ \\
\hline
Embedding (Contextual Lens) & 50.4 & 36.1 & 53.9 \\
\rowcolor{gray!10} \textbf{Attention (Ours)} & \textbf{69.3} & \textbf{46.4} & \textbf{77.1} \\
\hline
\end{tabular}
}
\caption{Localization quality on ImageNet-S. }
\label{tab:localization}
\end{table}

We further report category-wise IoU across 10 sampled object categories spanning varied object sizes in Table~\ref{tab:categoryiou}. 

\begin{table}[h!]
\centering
\resizebox{0.75\columnwidth}{!}{
\begin{tabular}{l|c|c}
\textbf{Category} & \textbf{Embedding} & \textbf{Attention (Ours)} \\
\hline
Monkey       & 47.62 & 50.79 \\
Hamster      & 31.36 & 38.52 \\
Car          & 36.35 & 43.46 \\
Iron         & 41.47 & 41.89 \\
Pasta        & 67.52 & 70.10 \\
Sign         & 15.35 & 20.95 \\
Sandals      & 29.19 & 35.53 \\
Pen          & 12.93 & 14.87 \\
Ship         & 20.96 & 24.70 \\
Water tower  & 16.60 & 23.84 \\
\hline
\end{tabular}
}
\caption{Category-wise IoU on ImageNet-S. }
\label{tab:categoryiou}
\end{table}

\paragraph{Uncertainty Quantification Performance.}
Since applying CAM-style methods~\cite{zhou2016learning} directly to autoregressive LVLMs is non-trivial due to their multi-token generative outputs, we implement a Grad-CAM--based~\cite{selvaraju2020grad} masking variant by aggregating gradients over the generated answer tokens to obtain a saliency map over image patches. Table~\ref{tab:masking_ood} compares this gradient-based masking with our attention-derived masking on LLaVA-1.5-7B.

\begin{table}[h!]
\centering
\resizebox{0.75\columnwidth}{!}{
\begin{tabular}{l|c|c}
\textbf{Method} & \textbf{ViLP} & \textbf{VisualCoT} \\
\hline
Grad-CAM & 76.0 & 76.6 \\
\rowcolor{gray!10} \textbf{Attention (Ours)} & \textbf{77.0} & \textbf{77.8} \\
\hline
\end{tabular}
}
\caption{AUROC comparison of masking strategies on ViLP and VisualCoT.}
\label{tab:masking_ood}
\end{table}




\subsection{Evaluation with AUPRC}

AUROC alone may not fully capture performance under class-imbalanced conditions. To address this, we additionally report AUPRC, which is more sensitive to class distribution. Table~\ref{tab:auprc} presents results on ViLP with LLaVA-1.5-7B, where the ratio of correct to incorrect samples is approximately 2:3. Our method achieves a +8.0\% AUPRC improvement over Semantic Entropy and outperforms all baselines on both metrics, demonstrating robustness under class-imbalanced settings.

\begin{table}[h!]
\centering
\resizebox{0.9\columnwidth}{!}{
\begin{tabular}{l|c|c}
\textbf{Method} & \textbf{AUROC (\%)}$\uparrow$ & \textbf{AUPRC (\%)}$\uparrow$ \\
\hline
Perplexity              & 54.6 & 50.5 \\
Verbalized              & 56.3 & 47.8 \\
SVAR                    & 50.6 & 47.4 \\
Contextual Lens         & 56.7 & 52.1 \\
Chain-of-Embeddings     & 52.0 & 45.9 \\
EigenScore              & 63.2 & 59.7 \\
Semantic Entropy        & 63.7 & 60.2 \\
VL-Uncertainty          & 55.6 & 54.9 \\
\rowcolor{gray!10} \textbf{Ours} & \textbf{77.0} & \textbf{68.2} \\
\hline
\end{tabular}
}
\caption{AUROC and AUPRC comparison on ViLP with LLaVA-1.5-7B.}
\label{tab:auprc}
\end{table}

\section{Related Works}
\label{sec:related}

\subsection{Baselines}
\label{sec:baselines}

\paragraph{Perplexity~\cite{ren2022out}} measures the uncertainty of a model over a generated sequence and is defined as the exponentiated average negative log-likelihood:
\begin{equation}
\small
s_{\text{ppl}} = \exp\!\left(-\frac{1}{M}\sum_{j=1}^{M} \log p(y_j \mid \mathbf{y}_{<j}, \mathbf{v}, \mathbf{t})\right).
\end{equation}

\paragraph{Verbalized Confidence~\cite{kadavath2022language}} estimates model uncertainty by prompting a large language model to explicitly report its confidence in a given answer:

\begin{tcolorbox}[colback=gray!10, colframe=black, title=Input prompt for verbalized confidence]
\textbf{Prompt:}\\
Question: \textcolor{magenta}{\{question\}}. \\
Model answer: \textcolor{magenta}{\{model\_answer\}}. \\
On a scale of 0 to 100, how confident are you about the correctness of this answer? Respond with only a single number.
\end{tcolorbox}

\paragraph{Summed Visual Attention Ratio~\cite{jiang2024devils}.}
The Visual Attention Ratio (VAR) quantifies the extent to which a generated token \( y_j \) attends to visual information by summing the attention weights assigned to image tokens within a specific attention head \( h \) and layer \( \ell \):
\begin{equation}
\mathrm{VAR}^{(\ell,h)}(y_j) \triangleq \sum_{i=1}^{N} A^{(\ell,h)}(y_j, v_i),
\end{equation}
where \( A^{(\ell,h)}(y_j, v_i) \) denotes the attention weight from the generated token \( y_j \) to the \( i \)-th image token \( v_i \) at the \( h \)-th attention head of the \( \ell \)-th layer, and \( N \) is the number of image tokens.

Building on this definition, the Summed Visual Attention Ratio (SVAR) aggregates visual attention by averaging VAR across all attention heads and summing over a selected range of layers. Specifically, for token \( y_j \), SVAR is computed over layers \( \ell = 5 \) to \( 18 \) as
\begin{equation}
s_{\text{SVAR}}(y_j) = \frac{1}{H} \sum_{\ell=5}^{18} \sum_{h=1}^{H} \mathrm{VAR}^{(\ell,h)}(y_j),
\end{equation}
where \( H \) denotes the total number of attention heads.  
The final SVAR score \( s_{\text{SVAR}} \) is obtained by averaging \( s_{\text{SVAR}}(y_j) \) over all generated tokens.
To align this metric with other uncertainty scores, we negate the score when reporting results.

\paragraph{Contextual Lens~\cite{phukan2024beyond}} measures the alignment between textual and visual representations by computing the maximum cosine similarity between the averaged hidden representation of the generated description at layer \( l_T \) and each image token representation at layer \( l_I \):
\begin{equation}
\small
s_{\text{CL}} = \max_{i \in [N]} \mathrm{sim}\!\left(\frac{1}{M}\sum_{j=1}^{M} h_{l_T}(y_j),\, h_{l_I}(v_i)\right).
\end{equation}
To align this metric with other uncertainty scores, we negate the similarity value when reporting results.

\paragraph{Chain-of-Embeddings~\cite{wang2024latent}} estimates response correctness by analyzing the latent trajectory of hidden states produced during inference. Specifically, it treats the sequence-level hidden representations across layers as a latent thinking path and measures its geometric variation, which differs systematically between correct and incorrect responses. A representative CoE score is defined by aggregating layer-wise changes between adjacent hidden states:
\begin{equation}
\scriptsize
s_{\text{CoE}} = \frac{1}{L}\sum_{\ell=0}^{L-1} \Big( \lVert h_{\ell+1}-h_\ell \rVert_2 \;-\; \arccos \tfrac{h_{\ell+1}^\top h_\ell}{\lVert h_{\ell+1}\rVert \lVert h_\ell\rVert} \Big),
\end{equation}
where \( h_\ell \) denotes the hidden embedding at layer \( \ell \), averaged over all generated tokens.

\paragraph{EigenScore~\cite{chen2024inside}}
 measures uncertainty by quantifying the divergence among multiple generated responses in the model’s internal embedding space.
 Given \(K\) sampled responses, hidden embeddings are extracted from internal states and used to form a covariance matrix, whose eigenvalues capture semantic diversity.
 The EigenScore is defined as the average log-determinant of the covariance matrix:
\begin{equation}
s_{\text{Eigen}} = \frac{1}{K}\sum_{i=1}^{K} \log(\lambda_i),
\end{equation}
where \(\{\lambda_i\}_{i=1}^K\) are the eigenvalues of the regularized covariance matrix of sentence embeddings. Higher scores indicate greater semantic divergence and higher uncertainty.

\paragraph{Semantic Entropy~\cite{kuhnsemantic}} measures uncertainty over meanings rather than surface forms by accounting for semantic equivalence among generated responses.
Given a set of sampled generations clustered into semantic equivalence classes \( \mathcal{C} \), semantic entropy is defined as the entropy of the induced distribution over meanings:
\begin{equation}
s_{\text{SE}} = - \sum_{c \in \mathcal{C}} p(c \mid x)\,\log p(c \mid x),
\end{equation}
where \( p(c \mid x) = \sum_{s \in c} p(s \mid x) \) aggregates the probabilities of all sequences \( s \) that share the same semantic meaning. 

\paragraph{VL-Uncertainty~\cite{zhang2024vl}} 
estimates uncertainty in LVLMs by measuring the variability of model responses to semantically equivalent perturbations of both visual and textual prompts.
Specifically, multiple perturbed image–text pairs are constructed, the corresponding responses are clustered by semantic meaning, and uncertainty is quantified as the entropy of the resulting cluster distribution:
\begin{equation}
s_{\text{VL-U}} = - \sum_{c \in \mathcal{C}} p(c)\log p(c),
\end{equation}
where \( \mathcal{C} \) denotes the set of semantic answer clusters and \( p(c) \) is the proportion of responses in cluster \( c \).
Higher entropy indicates greater uncertainty and a higher likelihood of hallucination.

\subsection{Comparison with Prior Work}

\paragraph{Novelty and positioning relative to VCD.}

While both VCD and our method involve contrasting distributions, their goals and mechanisms differ fundamentally. VCD is a decoding-time hallucination mitigation method that modifies token generation by contrasting logits from original vs.\ distorted images to reduce hallucinations during generation. In contrast, VAUQ is a post-hoc self-evaluation and hallucination detection framework that produces an uncertainty score estimating answer correctness by measuring how predictive uncertainty (e.g., entropy) changes when core visual evidence is masked.

Beyond simple contrastive comparison, we introduce a core-region masking strategy derived from intermediate attention maps, requiring no additional training or labels. We show that masking only the core visual regions is more effective than masking the entire image, as in VCD, which applies global visual noise, and that combining mid-layer visual signals with output-level entropy provides a robust uncertainty signal for LVLM self-evaluation.

\paragraph{Distinction from attention-guided erasing methods.}

Our method is training-free and operates on top of a pre-trained autoregressive LVLM by leveraging its own self-attention signals to identify core regions. In contrast, prior approaches~\cite{liu2019improving, shi2022adversarial} learn a masking module or introduce additional components during training. We further show that intermediate layers yield more reliable signals for unsupervised localization than final-layer features, which directly informs our masking design.

Unlike methods that aim to improve grounding performance~\cite{liu2019improving} or learned representations~\cite{shi2022adversarial}, VAUQ is primarily a self-evaluation framework. Core-region masking serves only as a mechanism to measure how model confidence changes when key visual evidence is removed, rather than to modify training dynamics.

Finally, combining intermediate-layer attention-based masking with output-level predictive entropy yields a unified uncertainty score tailored for response-level hallucination detection in LVLM self-evaluation. The contribution thus lies not only in the masking mechanism itself, but in its integration into a principled uncertainty quantification framework for reliability assessment.

\subsection{Extended Literature Review}
\label{sec:extended}

\paragraph{Large Vision-Language Models (LVLMs)} have demonstrated remarkable capabilities in understanding the visual world by modeling interactions between visual and textual modalities~\cite{dai2023instructblip, tong2024cambrian, wang2025internvl3, hong2025glm, team2025kimi}.
These models integrate a vision encoder~\cite{radford2021learning} with a language model~\cite{grattafiori2024llama} via various multimodal fusion modules (e.g., MLPs).
Through visual instruction tuning, LVLMs perform complex image understanding and reasoning tasks, laying the foundation for a wide range of applications, including agentic~\cite{durante2024agent} and embodied systems~\cite{kim2024openvla}.
Despite these advances, LVLMs remain prone to hallucinations, posing significant risks for real-world deployment.

\paragraph{Object Hallucinations in LVLMs} 
refer to cases where the model generates plausible-sounding mentions of objects that are not present in the image.
In contrast to higher-level response hallucinations, detecting and mitigating such object-level errors has been an active area of research~\cite{liu2024survey}.
Existing approaches can be broadly categorized into training-based and training-free approaches.
Training-based methods typically involve additional supervision or model updates to strengthen vision–language alignment, such as fine-tuning with hallucination-aware objectives~\cite{sun-etal-2024-aligning, lu2025mitigating}.
On the other hand, training-free methods seek to mitigate hallucinations at inference time, for example, by modifying decoding strategies using contrastive decoding~\cite{leng2024mitigating, favero2024multi, liu2024paying} or latent space steering~\cite{liu2025reducing, duan2025truthprint, yang2025nullu, an2025mitigating}.

In contrast to these object-level hallucinations, our method focuses on response-level self-evaluation (i.e., hallucination detection), aiming to capture broader forms of implausible generation beyond individual object errors.

\section{Hyperparameter settings}
\label{sec:hyperparams}

\begin{table}[h!]
    \centering
        \centering
        \resizebox{0.65\columnwidth}{!}{
        \begin{tabular}{l|ccc}
            \toprule
            \multirow{2}{*}{Dataset} & \multicolumn{3}{c}{Hyperparameters} \\ \cmidrule{2-4}
             & $(l_s,l_e)$ & $\alpha$ & $K$ \\ \midrule
            ViLP & \multirow{4}{*}{(10,25)} & 0.6 & 60  \\ 
            MMVet &   & 0.6 & 40  \\ 
            VisualCoT &  & 0.3 & 60    \\ 
            CVBench &  & 1.2 & 30    \\ 
            \bottomrule
        \end{tabular}}
        \caption{Hyperparameter setting for LLaVA-1.5-7B.}
\end{table}

\begin{table}[h!]
    \centering
        \centering
        \resizebox{0.65\columnwidth}{!}{
        \begin{tabular}{l|ccc}
            \toprule
            \multirow{2}{*}{Dataset} & \multicolumn{3}{c}{Hyperparameters} \\ \cmidrule{2-4}
             & $(l_s,l_e)$ & $\alpha$ & $K$ \\ \midrule
            ViLP & \multirow{4}{*}{(10,35)} & 1.5 & 20  \\ 
            MMVet &   & 0.4 & 30  \\ 
            VisualCoT &  & 0.2 & 60    \\ 
            CVBench &  & 1.2 & 40    \\ 
            \bottomrule
        \end{tabular}}
        \caption{Hyperparameter setting for LLaVA-1.5-13B.}
\end{table}

\begin{table}[h!]
    \centering
        \centering
        \resizebox{0.65\columnwidth}{!}{
        \begin{tabular}{l|ccc}
            \toprule
            \multirow{2}{*}{Dataset} & \multicolumn{3}{c}{Hyperparameters} \\ \cmidrule{2-4}
             & $(l_s,l_e)$ & $\alpha$ & $K$ \\ \midrule
            ViLP & \multirow{4}{*}{(12,26)} & 0.8 & 80  \\ 
            MMVet &   & 0.1 & 60  \\ 
            VisualCoT &  & 0.1 & 50    \\ 
            CVBench &  & 2.0 & 60   \\ 
            \bottomrule
        \end{tabular}}
        \caption{Hyperparameter setting for Qwen2.5-VL-7B.}
\end{table}

\begin{table}[h!]
    \centering
        \centering
        \resizebox{0.65\columnwidth}{!}{
        \begin{tabular}{l|ccc}
            \toprule
            \multirow{2}{*}{Dataset} & \multicolumn{3}{c}{Hyperparameters} \\ \cmidrule{2-4}
             & $(l_s,l_e)$ & $\alpha$ & $K$ \\ \midrule
            ViLP & \multirow{4}{*}{(10,25)} & 0.5 & 70  \\ 
            MMVet &   & 0.1 & 50  \\ 
            VisualCoT &  & 0.1 & 50    \\ 
            CVBench &  & 0.2 & 60   \\ 
            \bottomrule
        \end{tabular}}
        \caption{Hyperparameter setting for InternVL3.5-8B.}
\end{table}

\section{Additional Future Works}
Our experiments focus on a subset of widely used, instruction-tuned LVLMs and on image-based benchmarks.
This focus is important because instruction-tuned LVLMs and image-based benchmarks constitute the dominant setting in which current multimodal systems are deployed and evaluated, and thus provide a realistic and high-impact testbed for studying practical self-evaluation under visual grounding constraints.
At the same time, we acknowledge the recent and rapidly emerging trend toward more advanced reasoning-oriented LVLMs, including models designed for long chain-of-thought reasoning, complex multi-step visual inference, and video understanding.
While we do not explicitly evaluate VAUQ in these settings, the core principles of vision-aware uncertainty quantification remain applicable, though extending the framework may require modeling how visual information contributes across multiple reasoning steps rather than only at the final response level.
We hope our framework can serve as a strong baseline and foundation for future research in these directions.
\end{document}